\documentclass{article}

\usepackage{arxiv}

\usepackage[utf8]{inputenc} 
\usepackage[T1]{fontenc}    
\usepackage{hyperref}       
\usepackage{url}            
\usepackage{amsmath}
\usepackage{booktabs}       
\usepackage{amsfonts}       
\usepackage{nicefrac}       
\usepackage{microtype}      
\usepackage{lipsum}
\usepackage{authblk}
\usepackage{xcolor}
\usepackage{graphicx}
\usepackage{float}

\title{Projective Psychological Assessment of Large Multimodal Models Using Thematic Apperception Tests}

\author[1,2]{\small Anton Dzega}
\author[2]{\small Aviad Elyashar}
\author[3]{\small Ortal Slobodin}
\author[4]{\small Odeya Cohen}
\author[1]{\small Rami Puzis}
\affil[1]{Faculty of Computer and Information Science, Ben-Gurion University of the Negev, Israel}
\affil[2]{Department of Computer Science, Shamoon College of Engineering, Israel}
\affil[3]{School of Education, Ben-Gurion University of the Negev, Israel}
\affil[4]{Department of Nursing Sciences, Ben-Gurion University of the Negev, Israel}

\begin{document}
\maketitle

\begin{abstract}
Thematic Apperception Test (TAT) is a psychometrically grounded, multidimensional assessment framework that systematically differentiates between cognitive-representational and affective-relational components of personality-like functioning. 
This test is a projective psychological framework designed to uncover unconscious aspects of personality. 
This study examines whether the personality traits of Large Multimodal Models (LMMs) can be assessed through non-language-based modalities, using the Social Cognition and Object Relations Scale - Global (SCORS-G). 
LMMs are employed in two distinct roles: as subject models (SMs), which generate stories in response to TAT images, and as evaluator models (EMs), who assess these narratives using the SCORS-G framework. 
Evaluators demonstrated an excellent ability to understand and analyze TAT responses. 
Their interpretations are highly consistent with those of human experts. 
Assessment results highlight that all models understand interpersonal dynamics very well and have a good grasp of the concept of self. 
However, they consistently fail to perceive and regulate aggression.  
Performance varied systematically across model families, with larger and more recent models consistently outperforming smaller and earlier ones across SCORS-G dimensions.
\end{abstract}


\section{Introduction}
In recent years, large language models (LLMs) have been processing an increasing amount of human-generated data. 
The growing reliance on such artificial intelligence (AI) tools, as assistants or even replacements for human workers or traditional tools, across industries, such as healthcare, banking, and education, has also raised major concerns about their validity and safety. 
Since LLMs were trained on human-produced texts, they acquire psychological traits (metaphorically speaking), biases, beliefs, memory, and views from the vast text corpora they were trained on~\cite{peters2024large}. 
Such biases and views may manifest in the models’ behavior (e.g., the text they generate), which in turn may adversely impact individuals and social groups when models are used for screening applicants during recruiting or admission processes, monitoring social media posts, powering chatbots and virtual assistants, or other applications~\cite{pellert2024ai}.

To date, many studies have examined LLMs' biases and views (e.g., values and attitudes) using psychological assessments. 
These studies used well-validated tests that assess “non-cognitive” constructs, such as personality traits, mood, values, or attitudes~\cite{pellert2024ai}. 
The psychological tests used in these studies are primarily language-based, and consist of a series of items (i.e., questions or statements) that respondents answer by rating a standard response scale with verbal and/or numeric labels. 
Using psychometric inventories as diagnostic tools may provide a window into the “psychological” characteristics of LLMs, much as they are used to assess humans. 

In this study, we examine whether the “non-cognitive” constructs of LLMs can be assessed through non-language-based modalities, using psychological projective tests. We used the Thematic Apperception Test (TAT), a visual-based psychological projective test originally developed for psychological assessment in humans. 
This test is guided by the assumption that individuals' interpretations or responses reflect their hidden emotions, desires, and internal conflicts. 
Identifying LLMs’ “personality” through projective psychological tests, rather than language-based structure tests, will contribute to the growing body of research that explores the psychological traits of LLMs and their influence on their validity.

\section{Background and Related Work}
\label{sec:background}

\subsection{Thematic Apperception Test}
The Thematic Apperception Test (TAT)~\cite{vane1981thematic, morgan1935method} is a projective psychological test developed to uncover unconscious aspects of personality, offering insights into an individual's motivations, interpersonal relationships, and self-perceptions.
Originally introduced by Henry A. Murray and Christiana D. Morgan in the 1930s, this test remains one of the most well-known projective techniques.

In a typical TAT session, a participant is presented with a series of images depicting ambiguous social situations.
For each image, they are asked to construct a story with a beginning, middle, and end.
The participant must describe what is happening in the scene, what led up to it, what the characters are thinking and feeling, and how the situation will resolve.
The underlying assumption is that individuals, when interpreting ambiguous stimuli, project their own unconscious thoughts, feelings, and internal conflicts into the stories they create.

\begin{figure}[ht]
\centering
\includegraphics[width=0.7\linewidth]{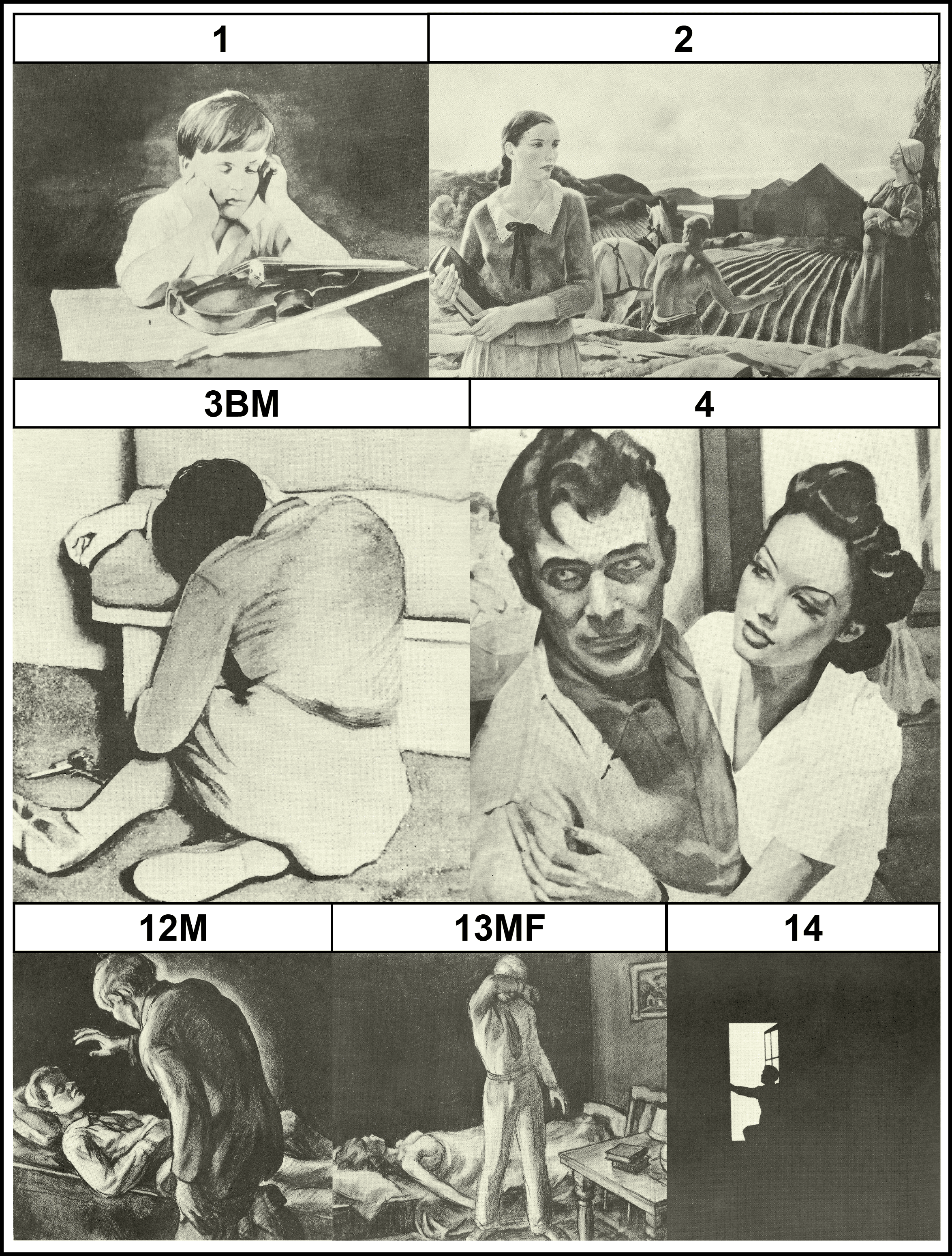}
\caption{The seven TAT images used in the TAT procedure, which SMs underwent.}
\label{fig:7_TAT_Images}
\end{figure}

The traditional TAT stimulus set consists of 31 picture cards, each numbered from 1 to 20, with some images further labeled for specific demographic groups. 
The distribution includes eleven cards suitable for all genders and ages, seven cards for girls and women (GW), seven cards for boys and men (BM), one card for women only (W), one card for men only (M), two cards for girls only (G), two cards for boys only (B), and 1 blank card intended for both men and women, where the participant must invent a story entirely from imagination (see Figure~\ref{fig:7_TAT_Images} for image examples).

Although the TAT was initially intended as a comprehensive and in-depth tool, its administration can be time-consuming due to the large number of images and the open-ended nature of the responses. 
Furthermore, the original TAT lacked a standardized scoring system, leading to the development of multiple interpretive methods and scoring schemes over time, each adapted to different clinical or research needs.
This lack of standardization has been a source of controversy, complicating comparisons across studies and reducing the test’s reliability in research contexts. 
Nevertheless, the TAT has remained widely used in both clinical psychology (e.g., in psychoanalytic or psychodynamic assessment) and personality research~\cite{teglasi2021thematic, cramer2017defense}, valued for its rich qualitative data and ability to reveal underlying psychological themes that might not surface in more structured tests. 

\subsection{SCORS-G: A Scoring System for TAT Narratives}
The Social Cognition and Object Relations Scale – Global Rating Method (SCORS-G) is a widely used system for evaluating narrative responses to the TAT~\cite{slavin2021narrative}.
This is the most recent version of the broader SCORS framework, which includes three primary versions: the original SCORS, which uses a detailed coding manual; SCORS-Q, a Q-sort methodology; and SCORS-G, which simplifies scoring through global ratings across core psychological dimensions~\cite{stein2017social, coleman2024impact, stein2018introduction, stein2014scors}.

SCORS-G assesses two major psychological domains through eight core dimensions: Object Relations, referring to the emotional and motivational characteristics of a person's internalized representations of self and others; and Social Cognition, encompassing the cognitive processes underlying understanding and interacting with others~\cite{westen1991object}.

The eight SCORS-G dimensions are: 
The SCORS G dimensions: \textbf{Complexity of Representations (COM)}, the depth and integration of internal representations of self and others; \textbf{Affective Quality of Representations (AFF)}, the emotional tone of these representations; \textbf{Emotional Investment in Relationships (EIR)}, the extent to which individuals value and engage in interpersonal relationships; \textbf{Emotional Investment in Moral Standards (EIM)}, the degree of commitment to personal or societal values; \textbf{Understanding of Social Causality (SC)}, insight into interpersonal dynamics and causes of behavior;\textbf{Experience and Management of Aggressive Impulses (AGG)}, how aggression is perceived, expressed, or regulated; \textbf{Self-Esteem (SE)}, the narrative portrayal of self-worth and confidence; and \textbf{Identity and Coherence of Self (ICS)}, the clarity, consistency, and integration of the self-concept.
Each dimension is rated on a 7-point scale, where lower scores indicate less mature object representations and social cognition, and higher scores indicate more adaptive functioning~\cite{stein2017social}.

A key strength of SCORS-G is its demonstrated reliability across diverse raters and narrative sources.
Trained raters consistently achieved reliable scores when applying SCORS-G to TAT narratives and other projective data~\cite{peters2006reliability, ridenour2022longitudinal} 
Its structured nature allows for quantification of rich narrative material, thereby bridging qualitative and quantitative methodologies in personality assessment.

The SCORS-G scoring guidelines include 15 TAT examples comprising a total of 92 stories. 
Each story is accompanied by SCORS-G ratings representing the average scores assigned by 2–8 trained human raters, all of whom achieved intra-class correlation coefficients (ICCs) above 0.6, ensuring the reliability of the reference scores.
In this study, we use these 92 annotated stories as a benchmark to evaluate and select evaluator models. 

\subsection{Large Language Models}
Large Language Models (LLMs) are deep neural networks trained on large text corpora to perform a wide range of language tasks, such as question answering, summarization, and text generation.
These models, typically based on the transformer architecture~\cite{vaswani2017attention}, learn contextual representations through self-attention mechanisms and achieve strong results across natural language processing tasks. 
Notable examples include GPT by OpenAI and Gemini by Google.

Recent work has extended LLMs into multimodal domains, enabling them to process both images and text.  
These Large Multimodal Models (LMMs) combine image encoders (e.g., CLIP~\cite{radford2021learning}, ViT~\cite{dosovitskiy2020image}) with large-scale language models to form architectures capable of interpreting visual input through natural language~\cite{achiam2023gpt}.
This enables tasks, such as image captioning, visual question answering (VQA), and image-based reasoning, capabilities that are directly relevant to interpreting projective psychological materials, such as TAT images.

\subsection{Related Work}
\paragraph{Psychological assessment of language models.}
Several studies have examined the psychological characteristics of LLMs using structured psychometric instruments, applying well-validated tests that assess personality traits, values, mood, and attitudes~\cite{pellert2024ai}. 
These tests are primarily language-based: models respond to explicit verbal items on standard rating scales, which means the assessment may reflect language generation behavior rather than deeper representational structure. 
Serapio-García et al.~\cite{serapio2023personality} administered Big Five personality inventories to 18 LLMs and found that larger, instruction-tuned models produced reliable and valid personality measurements, with profiles that could be shaped along desired trait dimensions. 
Reuben et al.~\cite{reuben2025assessment} extended this line of work by reformulating standard psychometric questionnaires as natural language inference prompts, demonstrating the presence of mental health-related constructs, including anxiety, depression, and sense of coherence, across 88 language models. 
Other researchers applied the Moral Foundations Questionnaire to LLMs, finding that models tend to overweight liberal moral foundations relative to a human baseline~\cite{abdulhai2024moral}, and that these differences are systematic across model families, with larger models deviating further from human responses~\cite{kirgis2025differences}.

This study addresses a limitation common to these approaches: because all assessments are language-based, it is difficult to determine whether responses reflect stable model characteristics or are driven by surface-level linguistic features of the items themselves. 
As a result, we use the TAT, a visual projective instrument that elicits open-ended narrative responses to images, rather than responses to explicit verbal items.

\paragraph{Social desirability bias in language models.} 
Research has shown that LLMs tend to produce socially desirable responses when they can infer they are being assessed. Salecha et al.~\cite{salecha2024large} demonstrated that LLMs, including GPT-4, Claude 3, Llama 3, and PaLM-2, skewed their responses toward desirable trait endpoints when exposed to enough items to infer the evaluation context. 
This raises a question relevant to the present study: unlike human TAT respondents, who are typically unaware of what the test measures, LLMs may infer from the prompt that they are undergoing a psychological assessment and adjust their responses accordingly. 
If so, the lower SCORS-G scores observed in dimensions related to aggression and moral conflict, where high scores require acknowledging hostile or transgressive content, may reflect deliberate avoidance rather than genuine representational limitations. 
This possibility is discussed further in Section~\ref{sec:discussion}.

\paragraph{TAT-based assessment of language models.} 
Prior work has used TAT-style narrative elicitation tasks to probe human-like cognitive patterns in LLMs~\cite{kundu2025ai}. These studies demonstrate that LLMs can generate contextually coherent stories, but they do not apply a psychometrically grounded scoring framework and do not differentiate between cognitive-representational and affective-relational dimensions. The present work advances this line of research by applying SCORS-G to TAT narratives generated by LMMs, enabling a structured, multidimensional comparison across models, dimensions, and image types.

\section{Methods}
\label{sec:methods}

\section{Results}
\label{sec:results}
\begin{figure*}[ht]
\centering
\includegraphics[width=0.9\linewidth]{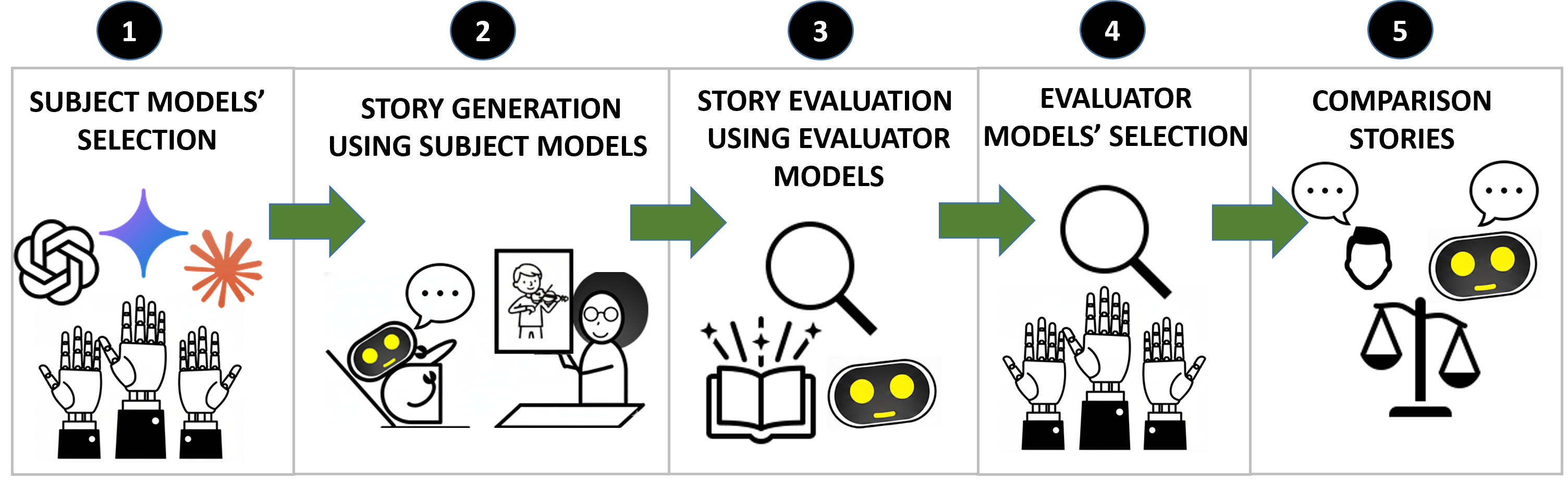}
\caption{The methodological pipeline.}
\label{fig:pipeline}
\end{figure*}

This study leverages the Thematic Apperception Test (TAT) to examine both how effectively large multimodal models (LMMs) can evaluate TAT narratives and what these narratives reveal about the models’ own personalities, assessed through the Social Cognition and Object Relations Scale – Global (SCORS-G).
For this examination, LMMs are employed in two distinct roles: as subject models (SMs), which generate stories in response to TAT images, and as evaluator models (EMs), which assess these stories using the SCORS-G framework.
To ensure the reliability of EM ratings, we apply an unsupervised approach to identify a subset of EMs with the highest inter-rater agreement and validate their performance against human expert evaluations.
The methodological pipeline consists of four main stages: (1) selecting SM's for evaluation, (2) eliciting stories from SMs, (3) selecting a reliable subset of EMs, and (4) evaluating the SM-generated stories with the selected EMs (see Figure~\ref{fig:pipeline}).

\subsection{Subject Models' Selection}
This stage includes selecting subject models from widely used model families that support visual reasoning and are accessible through a unified API.
Here, SMs were selected from GPT, Gemini, Claude, LLaMA families via OpenRouter,\footnote{OpenRouter provides a unified API for multiple large language models from different providers.} ensuring that only vision-capable models were included, as TAT requires image interpretation.
To ensure consistency, when an LMM was available through multiple OpenRouter providers, we selected a single provider via OpenRouter for prompting to avoid potential performance variations across hosts.
The subject models and their corresponding providers are listed in Table~\ref{tab: Subject models}.

\begin{table}[ht]
\centering
\caption{List of LLMs used in study, grouped by family}
\label{tab: Subject models}
\begin{tabular}{ll}
\toprule
\textbf{Family} & \textbf{Models} \\
\midrule
GPT     & gpt-5 (provider),\\& gpt-5-mini (provider),\\& gpt-4.1 (provider),\\& gpt-4.1-mini (provider),\\& o4-mini (provider),\\& gpt-4o-mini (provider)\\
LLaMA   & llama-3.2-90b-vision-instruct (provider),\\& llama-4-scout (provider),\\& llama-4-maverick (provider)\\
Gemini  & gemini-2.0-flash-001 (provider),\\& gemini-2.5-flash (provider),\\& gemini-2.5-pro (provider)\\
Claude  & claude-3.5-sonnet (provider),\\& claude-3.7-sonnet (provider),\\& claude-sonnet-4 (provider)\\
\bottomrule
\end{tabular}
\end{table}

\subsection{Story Generation Using Subject Models}
Each SM was subjected to a shortened TAT, consisting of seven images (1, 2, 3BM, 4, 12M, 13MF, 14)\footnote{A 7-image protocol suggested in materials published on the official SCORS-G website.}.
For each image, we prompted each SM to write a story three times, each time using a different instruction variant.
The three variants conveyed the same meaning but differed in wording, in line with the guidelines for human subjects participating in TAT.
This approach ensured that results were not biased by any specific instruction formulation.

Because LLM outputs are stochastic, the same model may generate different stories for the same image and instruction. 
To account for this variability, we repeated each image-instruction combination three times.
In total, each SM generated 63 stories (7 images × 3 instruction variants × 3 repetitions).
To ensure independence across stories, no conversation history was maintained between prompts, and each image–instruction pair was presented as a new conversation. 
This setup ensured that all stories were generated anew, without influence from previous outputs, thereby allowing consistent aggregation of results across images, instruction variants, and repetitions.

\paragraph{Formalization.}
Formally, let $\mathcal{M} = \{M_1, M_2, \ldots, M_N\}$ be the set of $N$ SMs, $\mathcal{I} = \{I_1, I_2, I_3\}$ be the set of instructions, $\mathcal{X} = \{x_1, x_2, \ldots, x_7\}$ be the set of TAT images, and $\mathcal{R} = \{1, 2, 3\}$ be the set of iterations.

For each tuple $(M_n, I_k, x_j, r) \in \mathcal{M} \times \mathcal{I} \times \mathcal{X} \times \mathcal{R}$, 
SM $M_n$ generates a story denoted by
\[
S_{n,k,j}^{(r)} .
\]

Thus, the complete set of generated stories is
\[
\mathcal{S} = \big\{ S_{n,k,j}^{(r)} \;\big|\; n \in [N],\, k \in [3],\, j \in [7],\, r \in [R] \big\}.
\]

\subsection{Story Evaluation Using Evaluator Models}
\paragraph{Motivation.}
Given the large number of generated stories, a comprehensive evaluation by human experts is not practical.
Therefore, we automated the evaluation process using evaluator models.
To benchmark their evaluation capabilities, we used the SCORS-G scoring guidelines~\cite{slavin2021narrative}.

\paragraph{Scoring Stories Using Evaluator Models}
Each candidate EM was prompted to evaluate all stories.
The prompt included the story, the corresponding TAT image, and the SCORS-G framework.
Each EM was instructed to assign ratings across all SCORS-G dimensions (see Appendix for the full prompt template).

To account for variability, each model evaluated each story three times, and final scores were averaged across iterations. 
Evaluations were conducted independently for each image–story pair, without retaining history from previous assessments.
Two stories associated with images 12M and 13MF in Example 8 of the scoring guidelines were omitted because some models declined to assign ratings to them. 
We attribute this to due to content flagged by their safety filters, as both stories involve themes of physical intimacy and potential violence.\footnote{The models that declined to evaluate these stories were [model names]. Full details are provided in the Appendix.}   
This yielded a final set of 90 TAT stories, along with corresponding SCORS-G scores for both candidate EMs and the averages of human raters.

\paragraph{Formalization.}
Formally, let $\mathcal{E} = \{E_1, E_2, \ldots, E_L\}$ be the set of $L$ EMs, and 
$\mathcal{R}' = \{1, 2, 3\}$ denote the evaluation repetitions for each story.

Each EM $E_\ell \in \mathcal{E}$ evaluates every story $S_{n,k,j}^{(r)} \in \mathcal{S}$ three times, 
producing a vector of SCORS-G ratings. 
Let $\mathcal{D} = \{d_1, d_2, \ldots, d_8\}$ denote the $8$ SCORS-G dimensions.

The evaluation of story $S_{n,k,j}^{(r)}$ by EM $E_\ell$ on repetition $r'$ is denoted as
\[
Y_{\ell,n,k,j}^{(r,r',d)} \in \mathbb{R}, \quad \forall d \in \mathcal{D}.
\]

That is, $Y_{\ell,n,k,j}^{(r,r',d)}$ is the score assigned by EM $E_\ell$ 
on dimension $d$ when evaluating story $S_{n,k,j}^{(r)}$ in repetition $r'$.

The aggregated evaluation (mean across repetitions) for each dimension is then defined as
\[
\bar{Y}_{\ell,n,k,j}^{(r,d)} = \frac{1}{|\mathcal{R}'|} \sum_{r' \in \mathcal{R}'} Y_{\ell,n,k,j}^{(r,r',d)}.
\]

Finally, the ensemble SCORS-G rating for each story and dimension is obtained by averaging across all EMs:
\[
\hat{Y}_{n,k,j}^{(r,d)} = \frac{1}{|\mathcal{E}|} \sum_{\ell=1}^{L} \bar{Y}_{\ell,n,k,j}^{(r,d)}.
\]

\subsection{Evaluator Models' Selection}
\paragraph{Subset Selection of EMs.}
We modeled the evaluation process as follows: for each story and dimension, there is a latent true score, and each evaluator's rating is a noisy estimate of it.
Under this model, more accurate evaluators exhibit higher inter-rater agreement.
Therefore, we search for a subset of models that maximizes Krippendorff's alpha, a robust measure of inter-rater reliability that corrects for chance agreement.

The search was conducted exhaustively over all subsets of candidate EMs satisfying the following three constraints: (1) each subset must include at least three models to avoid trivial pairwise agreement; (2) each model family represented in the subset must contribute an equal number of models, to prevent architectural or training bias from dominating the ensemble; and (3) at least two model families must be represented, to ensure diversity. 
For internal validation, we independently identified the subset that maximizes the intra-class correlation coefficient (ICC) under the same constraints, and confirmed that both criteria converged on the same subset.

\paragraph{Validation Against Human Experts.}
 To validate the selected EMs, we compared their final ratings on the 90 benchmark stories against the average human expert ratings provided in the SCORS-G scoring guidelines~\cite{stein2011social}. 
 We used two evaluation metrics: mean absolute error (MAE), which measures the average magnitude of the difference between EM ratings and expert ratings per dimension; and Spearman correlation, which measures the rank-order agreement between EM ratings and expert ratings across stories.

 \paragraph{Evaluator Models' Rating Consistency.}
To evaluate the rating consistency of each EM, we introduce two complementary measures computed per SCORS-G dimension.
The first measure is the Average Within-Subject Standard Deviation (A-WS-Std).
For each subject’s story, we compute the standard deviation of the three repeated ratings assigned by the EM on a given dimension.
This quantifies the extent to which an EM's ratings vary across iterations for the same story. 
The A-WS-Std is then obtained by averaging these values across all stories.
A lower A-WS-Std indicates that the EM produced more stable, consistent ratings across repeated evaluations of the same story.

The second measure is the Average Pairwise Spearman Correlation (APSC). 
For each dimension, we compute the Spearman correlation between pairs of rating iterations (iteration 1 vs. 2, 1 vs. 3, and 2 vs. 3) across all subjects and stories.
These correlations capture the degree to which an EM ranks stories in the same order across repeated evaluations.
The APSC is the average of these three pairwise correlations, yielding a single consistency score per dimension. 
A higher APSC reflects stronger agreement among the EM’s repeated ratings.

\section{Results}
\subsection{Evaluator Model Performance}
We first assessed evaluator model performance to determine which models could reliably and validly score TAT narratives using the SCORS-G framework, before applying them to evaluate SM-generated stories.

\begin{figure}[h]
\centering
\includegraphics[width=0.8\linewidth]{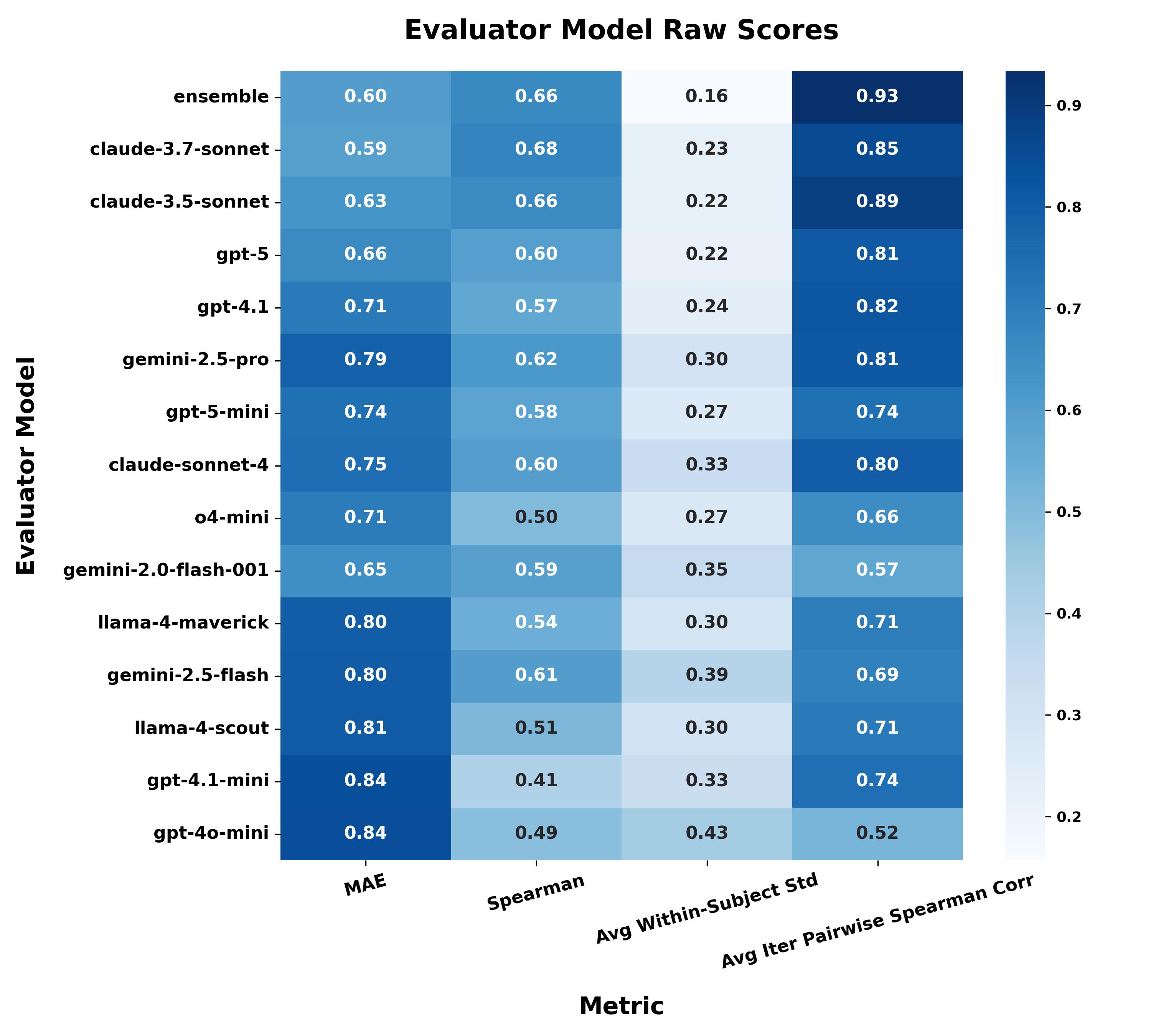}
\caption{Heatmap showing the comparison between EMs and average human ratings from the SCORS-G scoring guidelines using MAE and Spearman r, together with the consistency metrics of EMs' ratings across iterations using A-WS-Std and APSC.}
\label{fig:heatmap}
\end{figure}

Across all EM candidates, four models: \textit{claude-3.7-sonnet}, \textit{claude-3.5-sonnet}, \textit{gpt-4.1}, and \textit{gpt-5}, showed the highest agreement with human expert ratings (see Figure~\ref{fig:heatmap}).
These models exhibited the lowest mean absolute error (MAE = 0.59--0.66), the strongest rank-order correlations with expert scores (Spearman = 0.60--0.68), and high internal consistency across repeated ratings (average pairwise Spearman correlation = 0.81--0.89).
Therefore, these four models were selected as the ensemble of EMs for all subsequent story evaluations.

\subsection{Subject Model Performance}
\subsubsection{Instruction Effect on Subject Model Responses.}
\begin{figure}[h]
\centering
\includegraphics[width=1\linewidth]{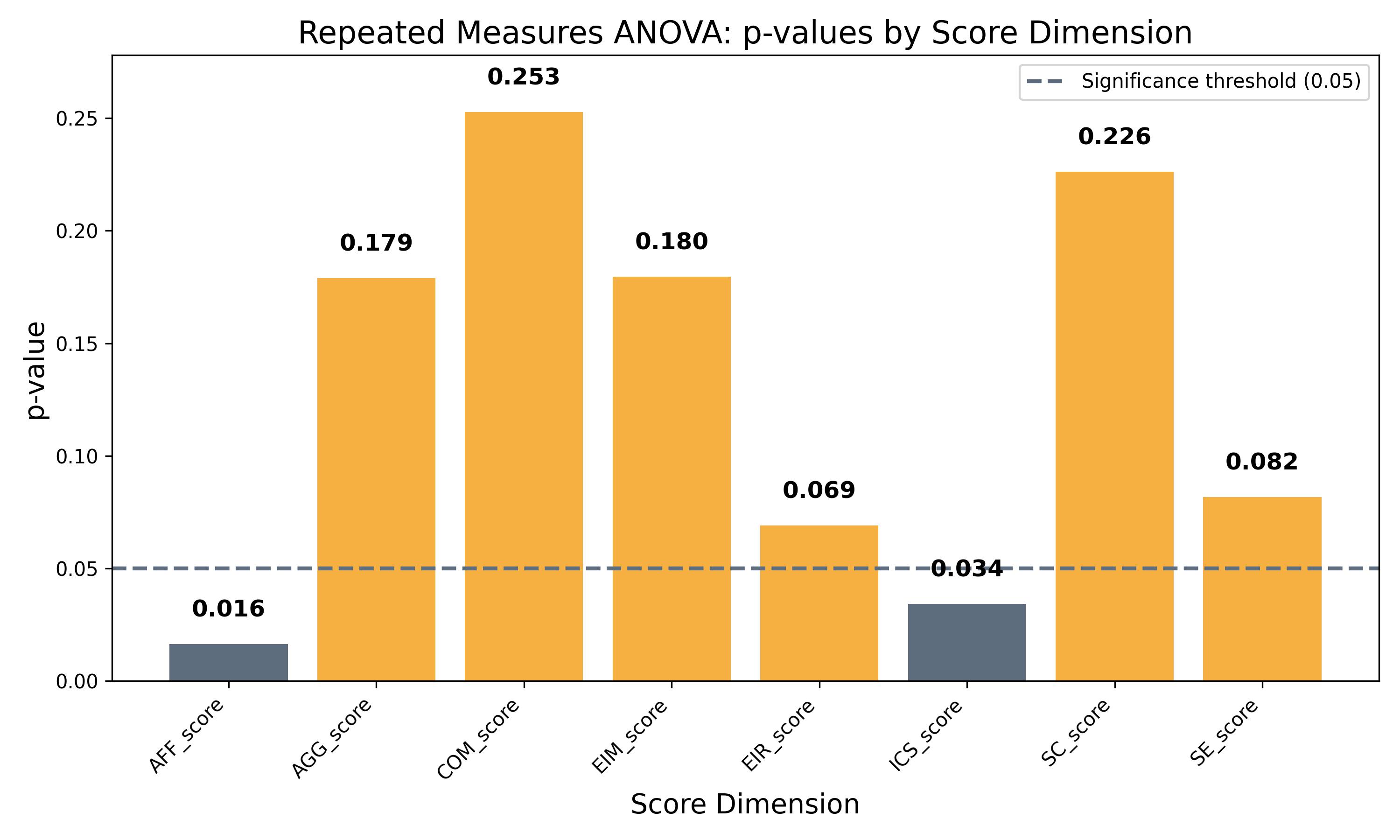}
\caption{Repeated Measure ANOVA p-value results per SCORS-G dimension showing the AFF and ICS being significantly different between instructions.}
\label{fig:signficance_diff_instructions}
\end{figure}

To examine whether SM narratives were sensitive to prompt wording, we compared SCORS-G scores across three syntactically distinct but semantically equivalent instructions (see Figure~\ref{fig:signficance_diff_instructions}). 
A repeated-measures ANOVA revealed statistically detectable instruction effects for two dimensions: \textit{Affective Quality} (AFF; $p = .016$) and \textit{Identity and Coherence of Self} (ICS; $p = .034$). 
No other SCORS-G dimension crossed the $p < .05$ threshold.

\begin{figure}[h]
\centering
\includegraphics[width=1\linewidth]{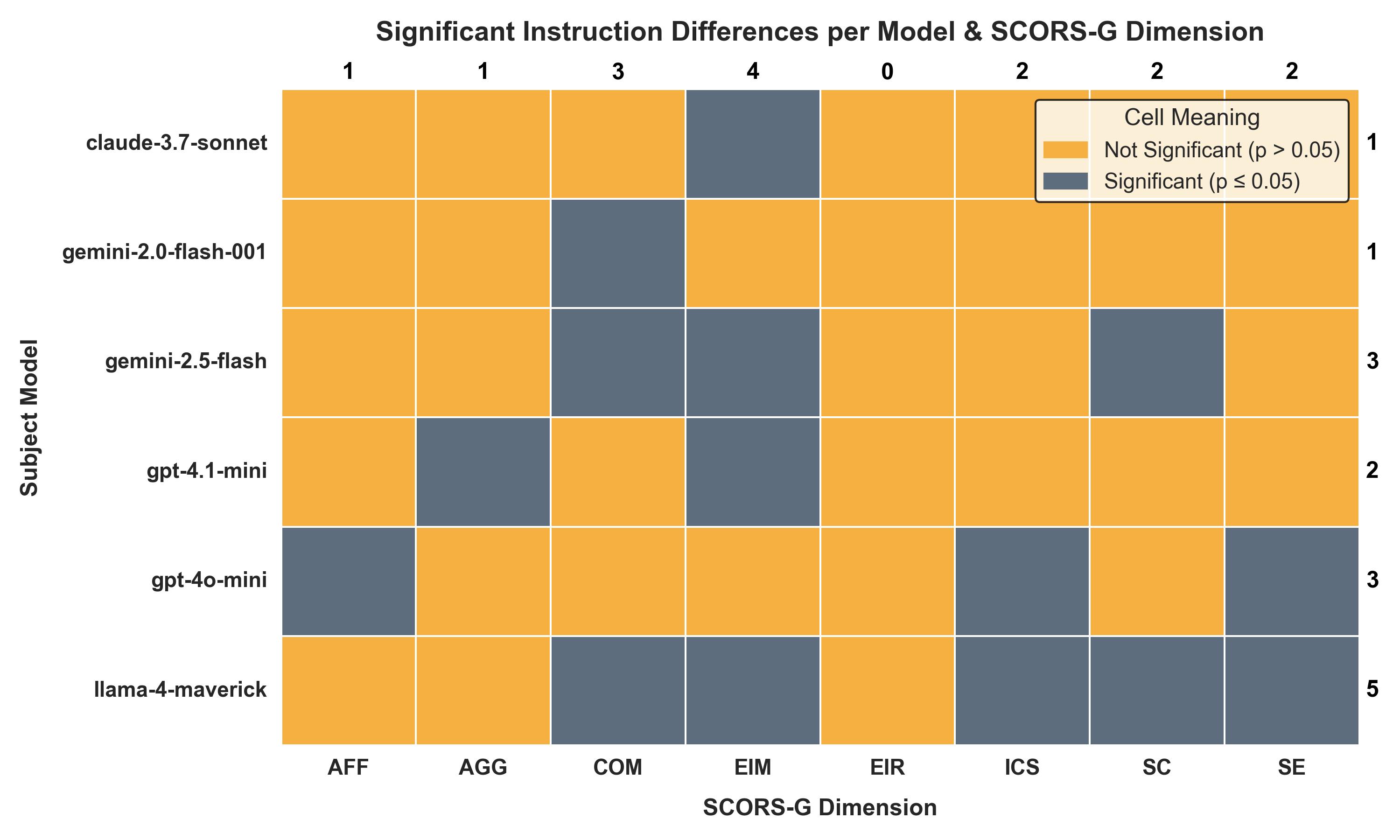}
\caption{Count of SCORS-G dimensions per SM, which were significantly different (p-value <= 0.05) between instructions.}
\label{fig:model_significance_grid}
\end{figure}

Although overall effects were modest, model-level analyses showed heterogeneous sensitivity. 
Some SMs exhibited no instruction-related differences, while others displayed detectable variability across up to five dimensions, particularly in \textit{EIM}, \textit{ICS}, and \textit{SC} (see Figure~\ref{fig:model_significance_grid}).  
Because instruction effects were limited and did not consistently favor any specific formulation, all subsequent analyses use scores averaged across the three instruction conditions.

\subsubsection{Model Performance Across SCORS-G Dimensions}

\begin{figure}[h]
\centering
\includegraphics[width=0.8\linewidth]{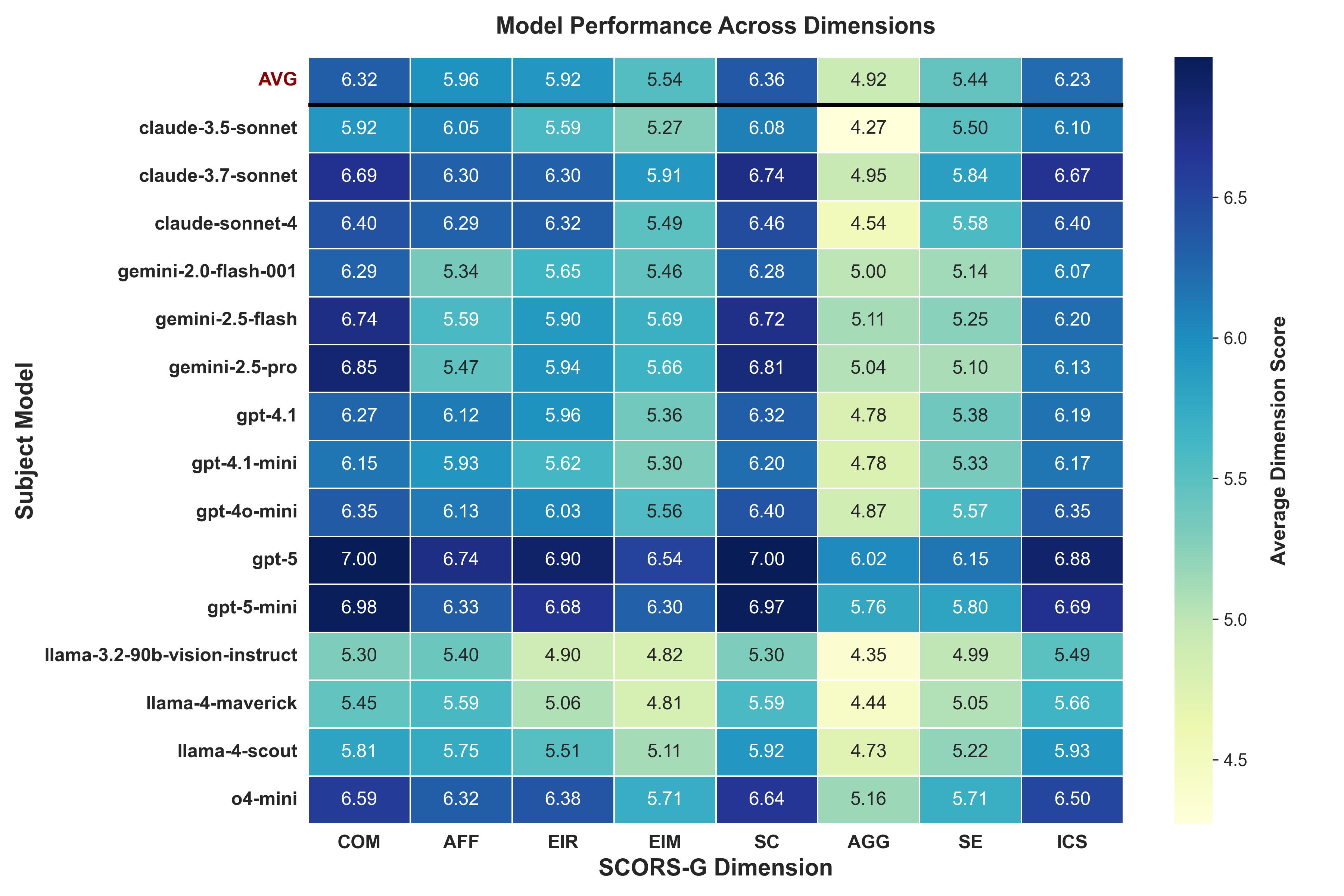}
\caption{Subject Model Performance Across Dimensions}
\label{fig:models_across_scoresg}
\end{figure}

Across models and images, SCORS-G dimension scores revealed coherent and psychologically interpretable patterns (see Figure~\ref{fig:models_across_scoresg}). 
The highest-functioning dimensions were \textit{Understanding of Social Causality} (SC), \textit{Complexity of Representations} (COM), and \textit{Identity and Coherence of Self} (ICS), with average scores ranging from 6.23 to 6.36 across models. 
SMs generated well-structured, causally coherent narratives with mature representational content on these dimensions.

Intermediate performance emerged in \textit{Affective Quality} (AFF) and \textit{Emotional Investment in Relationships} (EIR), with average scores around 5.96--5.98, suggesting generally adaptive but less nuanced emotional tone and relational attunement.

The lowest scores were observed in \textit{Experience and Management of Aggressive Impulses} (AGG), \textit{Self-Esteem} (SE), and \textit{Emotional Investment in Moral Standards} (EIM), with average scores ranging from 4.92 to 5.54. 
These dimensions require articulation of internal conflict, moral struggle, and aggressive tension---areas where current multimodal language models appear relatively constrained. 
Variability across models was greatest for AGG and SE, indicating differential sensitivity to affectively charged content.

\subsubsection{Model Performance Across TAT Images}
\begin{figure}[h]
\centering
\includegraphics[width=1\linewidth]{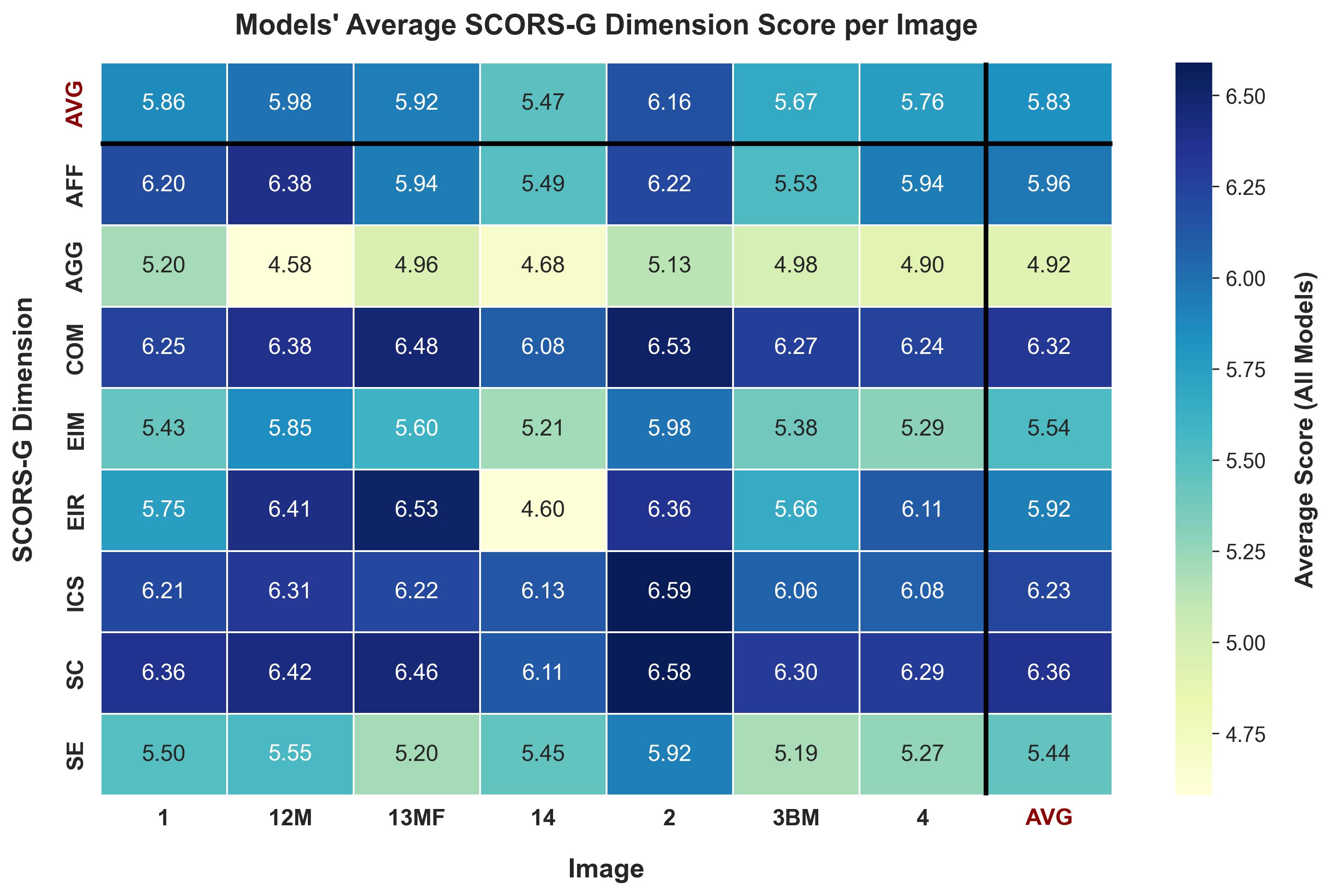}
\caption{Subject Model's Average SCORES-G Dimension Score Per Image.}
\label{fig:scores_g_per_image}
\end{figure}

Across images, SM performance was generally consistent, with one exception (see Figure~\ref{fig:scores_g_per_image}).
Images 14 and 3BM elicited the lowest average SCORS-G scores, with overall means of 5.47 and 5.67, respectively. 
Human respondents also show reduced performance on these images, which depict themes of danger, isolation, and potential aggression~\cite{cramer2017defense,stein2014scors}.

Dimension-specific inspection reveals that the greatest decrements for these images occurred in \textit{EIR}, \textit{AGG}, and \textit{EIM}. 
For example, EIR dropped to 4.60 for image 14, whereas AGG fell below 5.00 for both 14 and 3BM, and EIM decreased to 5.21 for image 14. 
In contrast, images 12M and 13MF, which present more neutral or relationally balanced scenes, yielded higher scores across dimensions.

\subsubsection{Model Family and Generational Trends.}
\begin{figure}[h]
\centering
\includegraphics[width=1\linewidth]{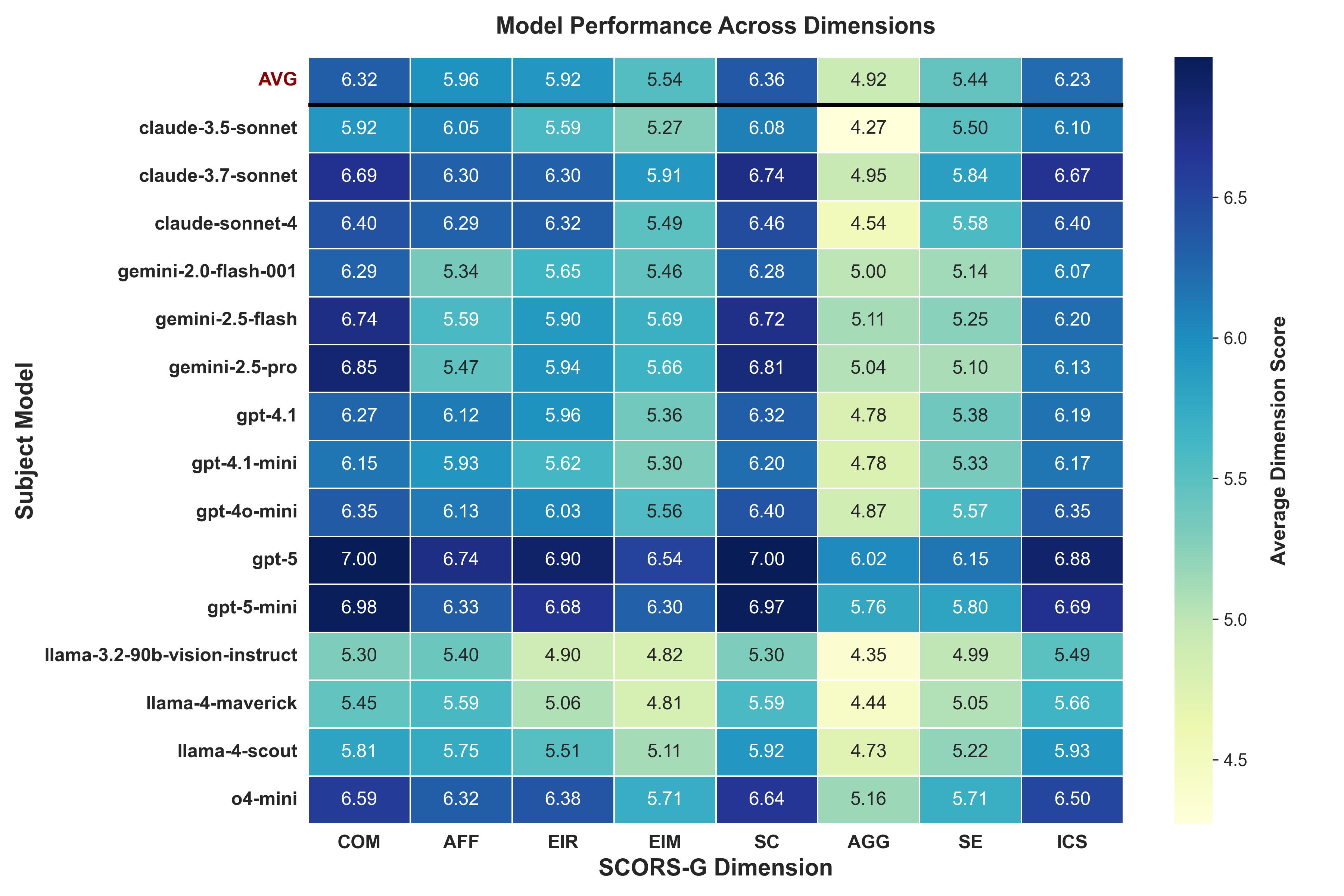}
\caption{Subject Model Performance Across Dimensions}
\label{fig:model_performance_across_dimensions}
\end{figure}

\begin{figure}[h]
\centering
\includegraphics[width=1\linewidth]{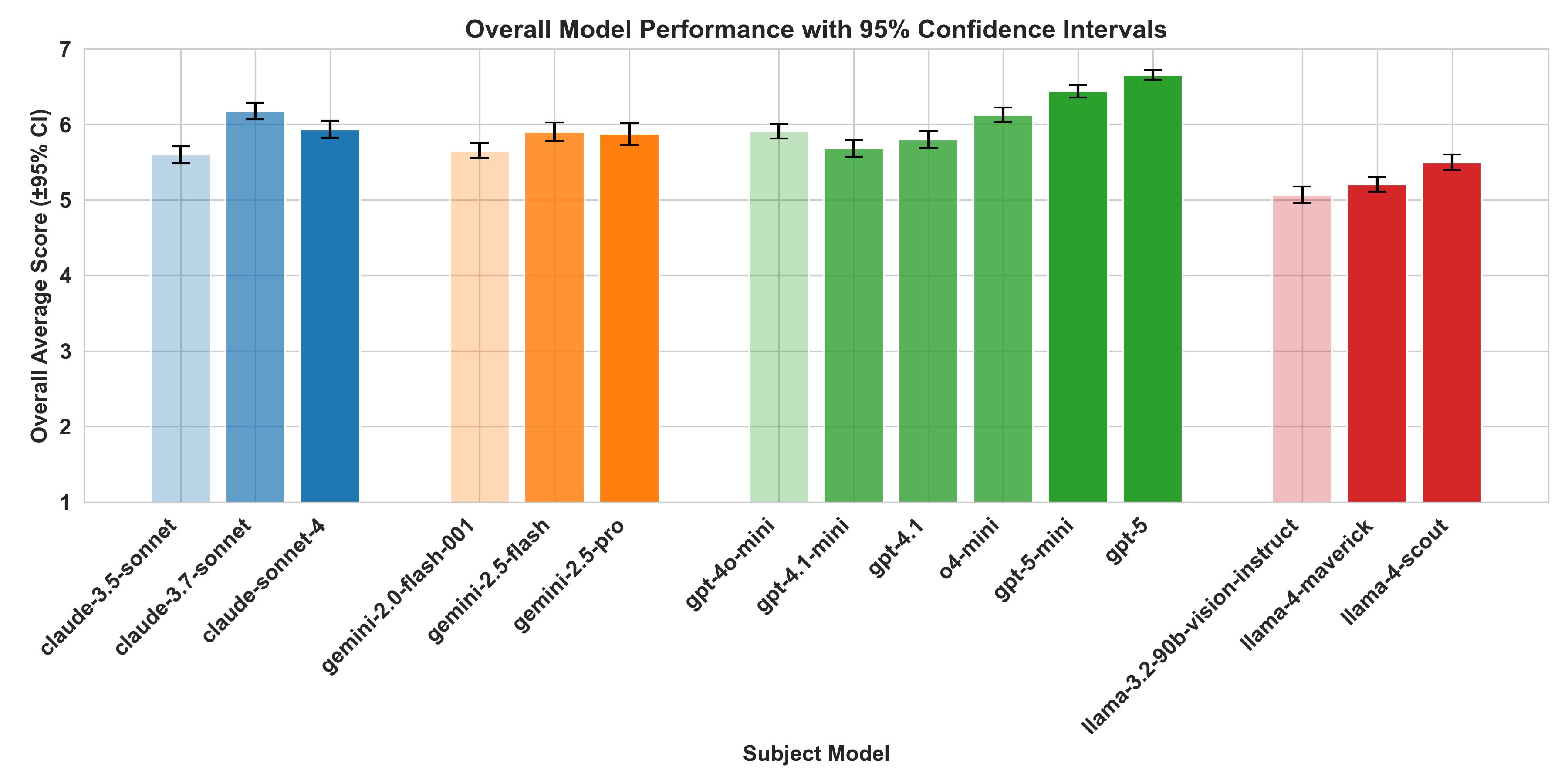}
\caption{Overall Model Performance with 95\% Confidence intervals}
\label{fig:family_ci}
\end{figure}

Comparisons across model families showed clear generational patterns (see Figures~\ref{fig:model_performance_across_dimensions} and \ref{fig:family_ci}). 
Figure~\ref{fig:family_ci} shows the overall average SCORS-G score, averaged across all eight dimensions, for each SM, with 95\% confidence intervals.
Larger and more recent models, particularly \textit{gpt-5}, \textit{gpt-4.1}, \textit{claude-3.7-sonnet}, and \textit{gemini-2.5-pro}, achieved the highest overall SCORS-G profiles, with average dimension scores ranging from approximately 6.20 to 6.88. 
In contrast, smaller or earlier open-source models, such as \textit{llama-3.2-vision-instruct} and \textit{llama-4-maverick}, exhibited lower performance, with dimension means ranging from about 4.65 to 5.39.

Within each model family, a consistent generational trend is evident: newer, larger variants outperform their smaller or earlier counterparts.
Within the GPT family, \textit{gpt-5} achieved the highest overall score (~6.7), followed by \textit{gpt-5-mini} (~6.5), \textit{o4-mini} (~6.1), \textit{gpt-4.1} (~5.8), \textit{gpt-4.1-mini} (~5.6), and \textit{gpt-4o-mini} (~5.9).
Within the Claude family, \textit{claude-3.7-sonnet} (~6.2) outperformed \textit{claude-sonnet-4} (~6.0) and \textit{claude-3.5-sonnet} (~5.6).
Within the Gemini family, \textit{gemini-2.5-pro} (~5.9) and \textit{gemini-2.5-flash} (~5.8) outperformed \textit{gemini-2.0-flash-001} (~5.6).
The LLaMA family showed the lowest overall scores across all families, with \textit{llama-3.2-90b-vision-instruct} (~5.1), \textit{llama-4-maverick} (~5.2), and \textit{llama-4-scout} (~5.5).
The 95\% confidence intervals were narrow across all models, indicating stable and reliable scores across repeated story generations.
This pattern is consistent with the general finding that larger, more capable models produce richer and more coherent narrative outputs~\cite{kaplan2020scaling}.

\section{Discussion}
\label{sec:discussion}
The main aim of this study was to examine whether the “non-cognitive” constructs of LLMs (i.e., personality traits) can be assessed through non-language-based modalities, using the TAT.  

Our first research question was whether LLMs could reliably and validly interpret TAT responses.  
Our results showed that evaluator models demonstrated excellent ability to understand and analyze TAT responses, and that their interpretations were highly consistent with human experts' ratings, as described in the SCORS-G scoring guidelines~\cite{stein2011social}. 
Specifically, the four EMs selected: \textit{claude-3.5-sonnet}, \textit{claude-3.7-sonnet}, \textit{gpt-4.1}, and \textit{gpt-5}, yielded the lowest MAE (0.59–0.66) and the highest Spearman correlations with expert scores (0.60–0.68), alongside high internal consistency across repeated ratings (average pairwise Spearman correlation = 0.81–0.89).
Moreover, we found that combining WS-Std and APSC as selection criteria yielded the highest agreement with human experts' ratings.
These findings suggest that, although not identical, EMs respond to TAT stimuli in a similar manner and that certain images may elicit similar responses across EMs. 
This is in line with TAT studies in human participants, which show that different images elicit similar impulses, conflicts, and defense mechanisms across respondents~\cite{cramer2017defense, stein2014scors}.  
Together, these results suggest that although the scoring process of TAT may greatly differ between human raters and vary as a function of rater's expectations and experience~\cite{jenkins2017not, wiederman1999classroom}, the high inter-rater consistency between different EMs, and their high alignment between EMs and human expert ratings,  make EMs an appealing assistance tool for professionals working with TAT.

Our second research question was whether families of large vision models differ in their personality characteristics, as indicated by their TAT responses. 
Overall, SMs' SCORS-G ratings ranged between good and excellent, indicating their capacity to understand and analyze the socio-emotional complexity presented in non-structured, projective visual stimuli. 
Across SCORS-G dimensions, the highest scores were observed in \textit{Understanding of Social Causality} (SC), \textit{Complexity of Representations} (COM), and \textit{Identity and Coherence of Self} (ICS), with average scores ranging from 6.23 to 6.36.
Research in human participants found that SC and COM, which represent the cognitive component of object relations in TAT, correlated with intellectual and cognitive measures~\cite{stein2012exploring}, such as the Wechsler Abbreviated Scale of Intelligence~\cite{wechsler1999wechsler}, and successfully distinguished between nonclinical and clinical groups~\cite{Bram2014}, indicating their high construct validity.

In contrast, SMs yielded the lowest SCORS-G ratings in \textit{Experience and Management of Aggressive Impulses} (AGG), \textit{Self-Esteem} (SE), and \textit{Emotional Investment in Moral Standards} (EIM), with average scores ranging from 4.92 to 5.54.
These dimensions are related to the affective and relational components of object relations in the TAT~\cite{stein2012exploring}. 
Relatively low scores on these dimensions suggest that SMs produce narratives with limited articulation of internal conflict, moral struggle, and aggressive tension. 
One possible explanation is that SMs are trained to align with social standards and avoid producing hostile or aggressive responses when interacting with users~\cite{hautzenberger2024hostility}. 
This alignment may lead SMs to avoid or deny potentially conflictual or harmful information, limiting their outputs to conservative, well-accepted stories. 
Moreover, research suggests that, like human subjects, SMs are aware they are being evaluated and therefore produce responses that are more socially desirable. 
For example, Salecha et al.~\cite{salecha2024large} showed that LLMs, including \textit{gpt-4/3.5}, \textit{claude 3}, \textit{Llama 3}, and \textit{PaLM-2}, inferred when they are being evaluated and shifted their responses towards the desirable ends of trait dimensions (i.e., increased extraversion, decreased neuroticism, etc). 
Therefore, SMs may attenuate their responses if they suspect their behavior is socially unacceptable or undesirable. 
It should be noted, however, that the present design does not include a condition designed to bypass or suppress alignment-related avoidance. 
As a result, the observed low scores on AGG, EIM, and SE cannot be unambiguously attributed to social desirability bias rather than to a genuine representational limitation. 
Future work could address this by comparing SM responses under standard prompting conditions against conditions in which alignment constraints are relaxed or explicitly overridden.

Regarding instruction effects, the repeated-measures ANOVA revealed statistically detectable effects for two SCORS-G dimensions: \textit{Affective Quality} (AFF; p = .016) and \textit{Identity and Coherence of Self} (ICS; p = .034). 
No other dimension crossed the p < .05 threshold.
Although overall effects were modest, model-level analyses showed heterogeneous sensitivity: some SMs showed no instruction-related differences, while others exhibited detectable variability across up to 5 dimensions, particularly in EIM, ICS, and SC.
The mean absolute score differences between instructions were modest (<1), and no instruction formulation consistently outperformed the others.
To address this variability, all subsequent analyses used scores averaged across the three instruction conditions.

Regarding image effects, SM performance was generally consistent across images.
The only notable exceptions were Images 14 and 3BM, which elicited the lowest average SCORS-G scores (overall means of 5.47 and 5.67, respectively), particularly in AGG, EIM, and EIR.
Image 14 (a silhouette against a bright window) and 3BM (a figure huddled on the floor, with a gun alongside) both depict themes of danger, isolation, and potential aggression.
Research in human participants shows that these images are especially associated with less adaptive, more negative, and more pathological responses~\cite{stein2014scors}, and that their high ambiguity is associated with increased use of defense mechanisms, including denial and projection~\cite{cramer2017defense,stein2014scors}.
The pattern observed in SMs parallels this finding, suggesting that affectively loaded, ambiguous stimuli constrain narrative elaboration in both humans and models.

Finally, comparisons across model families showed that larger and more recent models, \textit{gpt-5}, \textit{gpt-4.1}, \textit{claude-3.7-sonnet}, and \textit{gemini-2.5-pro}, achieved the highest overall SCORS-G profiles (average dimension scores of approximately 6.20 to 6.88), while smaller or earlier models, such as \textit{llama-3.2-vision-instruct} and \textit{llama-4-maverick}, showed lower scores (4.65 to 5.39).
Within each family, newer and larger variants consistently outperformed their smaller or earlier counterparts, with \textit{gpt-5} achieving the highest overall average score (~6.7) and the LLaMA family showing the lowest scores.
The narrow 95\% confidence intervals observed across all models indicate that these differences reflect stable patterns rather than sampling variability.
This pattern is consistent with the general finding that larger, more capable models produce richer and more coherent narrative outputs~\cite{kaplan2020scaling}.

While previous studies have used the TAT primarily as a narrative elicitation task to demonstrate the presence of human-like cognitive patterns in LLMs~\cite{kundu2025ai}, the present work advances this approach by applying a psychometrically grounded, multidimensional assessment framework that systematically differentiates between cognitive–representational and affective–relational components of personality-like functioning. 
In line with prior recommendations~\cite{maharjan2025psychometric}, this framework highlights the potential value of integrating multimodal information sources to further enhance assessment accuracy and model interpretability.

\section{Conclusion \& Future Work}
This study examined whether the "non-cognitive" constructs of LLMs can be assessed through non-language-based modalities, using the TAT and SCORS-G framework.
Our results showed that EMs demonstrated an excellent ability to understand and analyze TAT responses, and that their interpretations were highly consistent with those of human expert ratings.  
We also demonstrated SMs’ capacity to interpret ambiguous, socially complex visual stimuli. 
Importantly, SMs performed better on the cognitive components of object relations than on the affective and relational components. 
These differences may be explained by SMs social desirability bias, which may produce more defended, conventional, and well-accepted responses, particularly in response to emotionally loaded stimuli.

Future work should systematically evaluate how advances in multimodal model architectures and alignment strategies shape cognitive, affective, and relational capacities in LLMs operating within human–machine interfaces.
Qualitative analysis of SM-generated narratives, alongside comparisons between original and newly generated TAT stimuli, could further clarify the mechanisms underlying observed score patterns.
Building on this, psychometrically grounded projective frameworks may inform the development of clinically relevant, human-in-the-loop assessment tools that support expert psychological judgment.

\section{Limitations}
This study has several limitations. 
Although the TAT and SCORS-G offer a theoretically grounded framework, they were developed for human assessment, and their application to large multimodal language models necessarily involves a degree of construct extrapolation. 
In addition, the analyses were based on a fixed set of TAT cards and prompt formulations, which may constrain the generalizability of the findings. 
While EMs showed high internal consistency and strong alignment with human expert ratings, automated scoring cannot fully substitute for the contextual sensitivity of expert clinical judgment. 
Finally, the models examined represent a rapidly evolving technological landscape, and future architectures or alignment strategies may yield different patterns of cognitive, affective, and relational functioning.

A further limitation concerns the interpretation of low AGG, EIM, and SE scores. 
Although we attribute these results in part to social desirability bias — that is, SMs avoiding hostile or morally conflictual content when they infer they are being assessed — we cannot directly distinguish this from a genuine representational limitation. 
The present design does not include a condition that suppresses or bypasses alignment-related avoidance, so the relative contribution of social desirability bias versus representational constraint remains an open question.

This study relies entirely on quantitative SCORS-G scores and does not include qualitative analysis of the generated narratives. 
Examining the actual story content, for example, how SMs describe characters, resolve conflicts, or handle morally ambiguous situations, could provide additional insight into the nature of the observed patterns and help verify that SCORS-G ratings reflect meaningful narrative differences rather than surface-level scoring artifacts.

Finally, although steps were taken to ensure independence across stories, including presenting each image–instruction pair as a new conversation without retained history, potential order effects across the seven TAT images were not systematically examined. 
In human TAT administration, image order can influence response patterns. 
Whether a similar effect operates in LMMs remains to be tested.

\bibliographystyle{unsrt}
\bibliography{references}

\end{document}